\documentclass{article}
\usepackage{spconf,amsmath,amssymb,graphicx,hyperref}

\title{Beyond Saliency: Enhancing Explanation of Speech Emotion Recognition with Expert-Referenced Acoustic Cues}
\name{Seham Nasr$^1$, Zhao Ren$^2$, David Johnson$^1$}
\address{$^1$Center for Cognitive Interaction Technology (CITEC), Bielefeld University, Germany\\
$^2$Cognitive Systems Lab, University of Bremen, Germany}

\usepackage{booktabs,threeparttable,siunitx}
\usepackage[table]{xcolor}
\definecolor{rowhighlight}{gray}{0.92}
\newcommand{\best}[1]{\cellcolor{rowhighlight}\textbf{#1}}
\usepackage{verbatim}
\usepackage{tabularx}
\usepackage{adjustbox}
\newcolumntype{L}{>{\raggedright\arraybackslash}X}
\usepackage{spconf,amsmath,graphicx,hyperref}
\usepackage{tabularx}
\usepackage{paralist}
\usepackage{setspace}
\usepackage{hyperref}
\usepackage{multirow}
\usepackage{diagbox}
\usepackage{geometry}
\begin{document}
%
\maketitle
\begin{abstract}
Explainable AI (XAI) for Speech Emotion Recognition (SER) is critical for building transparent, trustworthy models. Current saliency-based methods, adapted from vision, highlight spectrogram regions but fail to show whether these regions correspond to meaningful acoustic markers of emotion, limiting faithfulness and interpretability. We propose a framework that overcomes these limitations by quantifying the magnitudes of cues within salient regions. This clarifies \textit{``what"} is highlighted and connects it to \textit{``why"} it matters, linking saliency to expert-referenced acoustic cues of speech emotions. Experiments on benchmark SER datasets show that our approach improves explanation quality by explicitly linking salient regions to theory-driven speech emotions expert-referenced acoustics. Compared to standard saliency methods, it provides more understandable and plausible explanations of SER models, offering a foundational step towards trustworthy speech-based affective computing.
\end{abstract}

%
\section{Introduction}
\label{sec:intro}

\begin{figure}[ht!]
    \centering
    \includegraphics[width=\linewidth]{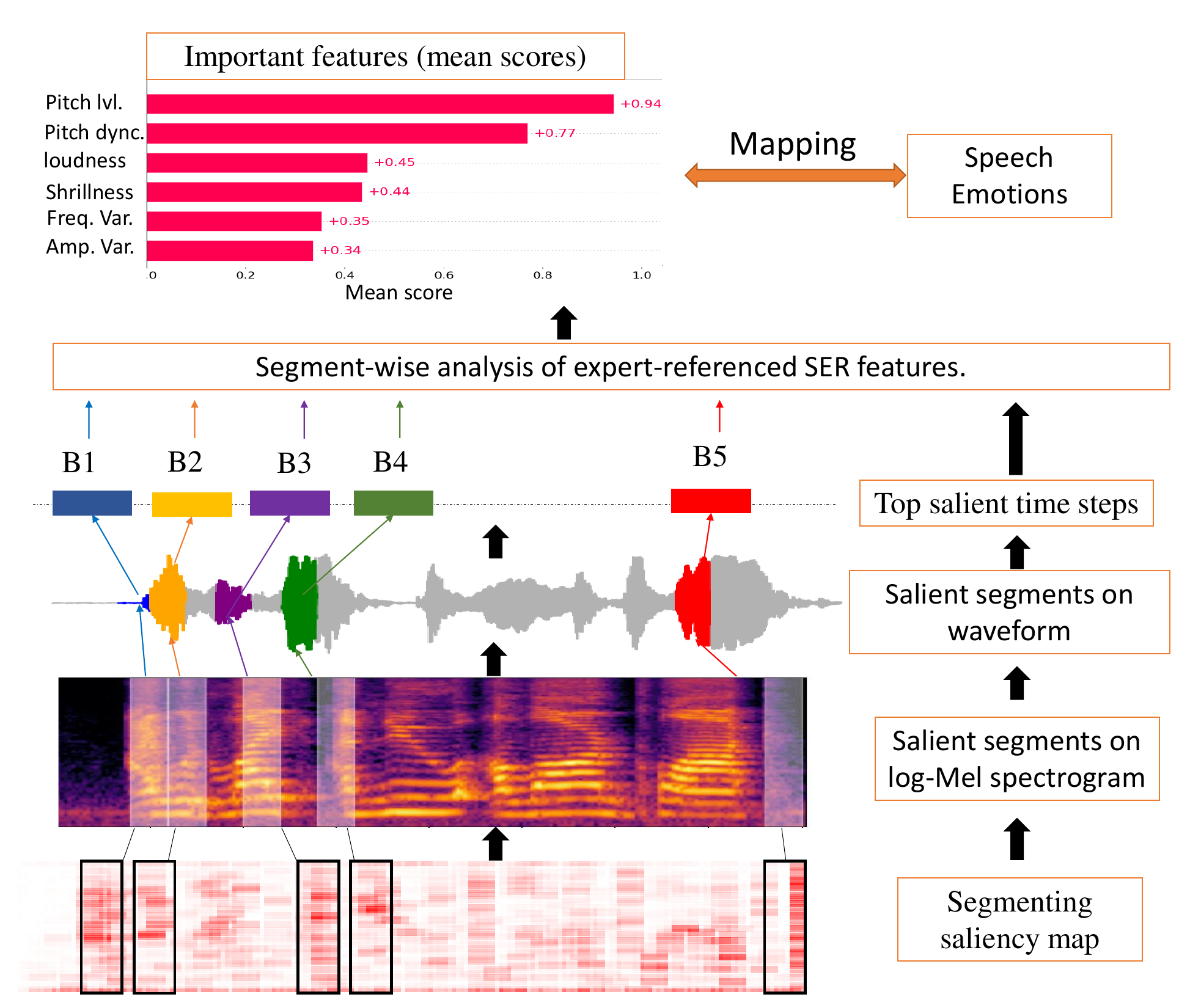}
    \caption{\textbf{The proposed framework.} Saliency maps are segmented (bottom), and top regions are projected onto the log-Mel spectrogram and waveform. The resulting temporal segments are analyzed for expert-referenced features, with \textit{``Mean''} scores mapped to speech emotions. }
    \label{fig:methodpipeline}
\end{figure}
With the rapid development of speech technologies across diverse applications~\cite{lestari2023deep}, speech emotion recognition (SER) has gained importance for monitoring mental health and enhancing well-being by detecting emotions from speech~\cite{sun2025weakly}. 
Recent studies have reported promising SER performance~\cite{zhao2019exploring}, but the opaque decision-making of deep neural networks raises concerns about reliability in sensitive domains such as healthcare, where misinterpretation can result in harmful interventions~\cite{ahmad2024mental}. These challenges show the need for explainable SER models that enhance transparency and trustworthiness.

Explainable AI (XAI) methods aim to improve transparency by providing insights into model behavior 
~\cite{jayasinghe2025systematic,Nasr2024IndMaskIE}. While XAI research has largely focused on vision and text~\cite{Johnson_2024}, XAI for speech is equally crucial for affective tasks~\cite{akman_audio_2024}. Recent SER studies have applied saliency-map XAI methods on spectrograms, such as occlusion sensitivity, and GradCAM~\cite{kim2024speech,becker2024audiomnist}.  

However, explaining speech models remains challenging because speech lacks an intuitive visual form. Current approaches often apply saliency maps on spectrograms \cite{kim2024speech, jayasinghe2025systematic}, which are difficult for non-experts to interpret and risk oversimplification by merely highlighting regions without contextual meaning~\cite{zhang2022towards,usha2024advanced,kim2024speech}. To our knowledge, methods for augmenting saliency maps in SER have not yet been explored, indicating a need for methods to enhance saliency map explanations with \textit{understandability} (what are the important features of SER models), \textit{fidelity} (how well do explanations reflect model reasoning), and \textit{plausibility} (how well do saliency-based XAI methods unveil the model irrationality during model misclassifications)~\cite{tomsett2020sanity}.

To address these issues, we propose a framework that quantifies acoustic cues within the most important regions of saliency maps, grounded in expert-referenced speech emotion theory, and links them to emotional arousal levels. This approach clarifies \textit{``where''} salient regions occur in time, \textit{``why''} it matters for model reasoning by identifying key regions in spectrograms, and \textit{``what''} they represent in terms of emotion-relevant acoustics. Our main contributions are:
\begin{itemize}
    \item A novel framework (presented in Figure \ref{fig:methodpipeline}) for enhancing saliency-map explanations in SER, grounded in expert-referenced emotion theory and correlated with emotional arousal that is summarized in Table \ref{tab:vocal_cues_emotion}. 
    \item A systematic evaluation addressing two research questions: (1) How well do different XAI methods capture expert-referenced cues? (2) Do results generalize across datasets and emotions?
    \item We introduce two levels of baseline validation procedures demonstrating that the framework produces faithful and robust SER explanations.    
\end{itemize}


\begin{table*}[t]
\centering
\footnotesize
\setlength{\tabcolsep}{4pt}
\renewcommand{\arraystretch}{1.15}
\caption{Speech emotion patterns from expert-referenced acoustic cues~\cite{scherer2003vocal,jacob2016speech,eyben2015geneva,wiechmann2025challenges,zhang2022towards}.}
\begin{tabularx}{\textwidth}{@{} l l c c c L @{}}
\toprule
\textbf{Acoustic} & \textbf{Perceptual feature}&\textbf{Unit} & \textbf{High} & \textbf{Low} & \textbf{Definition} \\
\midrule
Loudness               & Loudness  &  [sones]         & Happy, Anger, Fear          & Neutral, Sad, Calm     & A loud, high–intensity voice. \\
High-Freq Energy & Shrillness         & [dB/kHz] & Happy, Anger, Fear          & Neutral, Sad, Calm     & Energy above 500 Hz, perceived as harsh. \\
Jitter            & Frequency variation&[$\%$]  & Happy, Anger, Fear          & Neutral, Sad, Calm     & Irregularity in freq. from cycle to cycle. \\
Shimmer           & Amplitude variation& [dB]  & Happy, Anger, Fear          & Neutral, Sad, Calm     & Irregularity in amp. from cycle to cycle. \\
F0 Mean           & Pitch level        &[semitones re. 27.5 Hz]  & Joy, Anger                  & Sadness, Calm          & Mean fundamental freq. in speech. \\
HNR            & Breathiness/hoarseness  & [dB]    & Joy, Anger                  & Boredom, Sadness       & Harmonics-to-noise ratio in dB. \\
\bottomrule
\end{tabularx}
\label{tab:vocal_cues_emotion}
\end{table*}


\section{Enhancing Saliency Maps}
\label{sec:method}


Our framework, outlined in Figure~\ref{fig:methodpipeline}, augments SER saliency explanations by linking decision-relevant opaque regions to expert-referenced acoustic cues (i.e., important features), quantifying their magnitudes, then mapping them to arousal levels. We frame acoustics at a perceptual level to enhance understandability. 

\begin{figure}[h!]
    \centering
    \includegraphics[width=\linewidth]{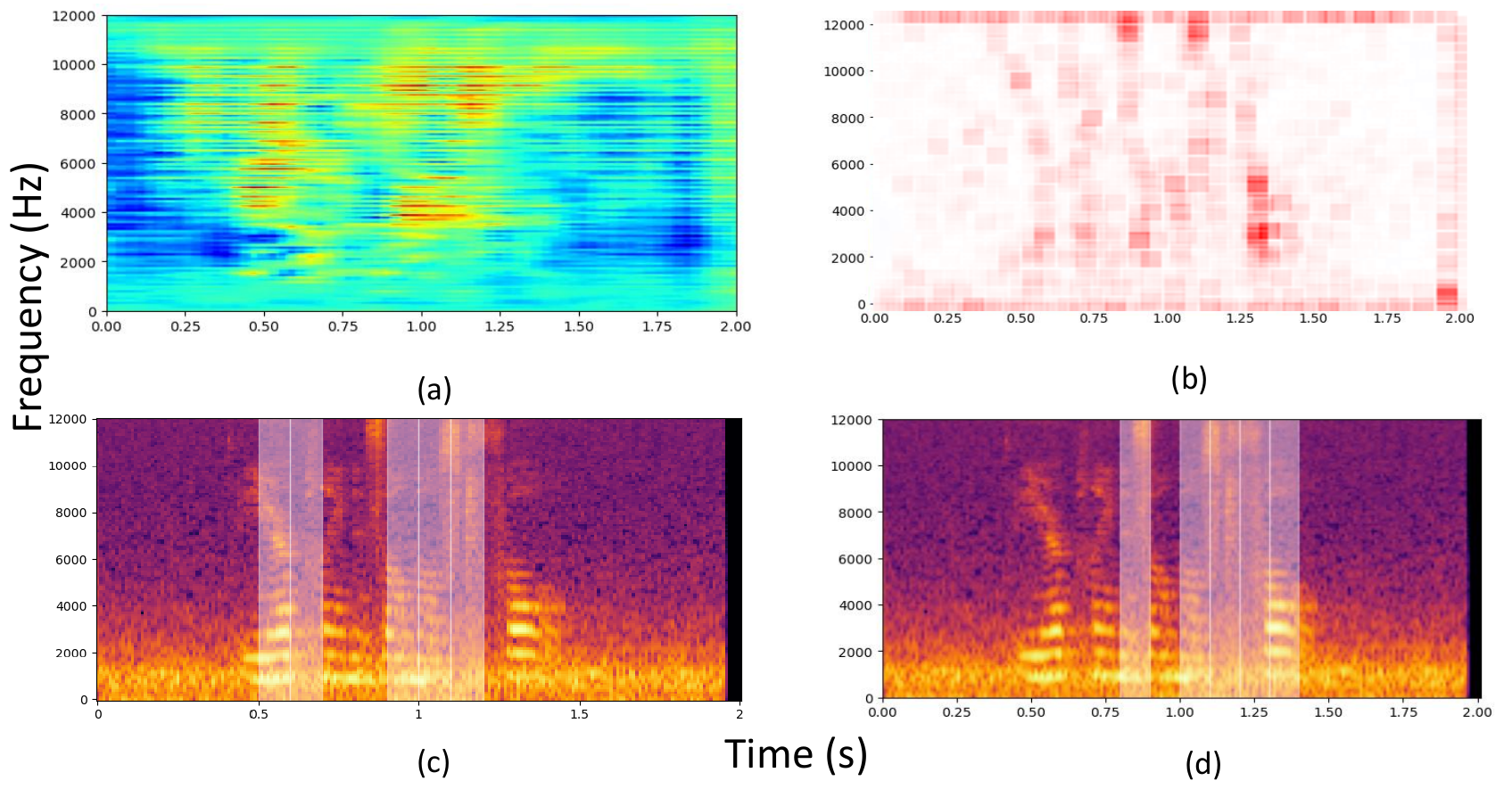}
    \vspace{-20pt}
    \caption{Samples instance-level saliency maps from XAI methods in our approach for emotion \textit{``sad''}. (a) OS XAI, (b) CRP XAI, and (c) and (d) highlight the salient regions projected on the input log-mel spectrogram.}
    \label{fig:saliencymaps}
\end{figure}

\noindent\textbf{Saliency Maps:} In SER, the input log-Mel spectrogram is a tensor $X \in \mathbb{R}^{H \times W \times 1}$, where ${H}=$ number of frequency bins, ${W}=$ number of time frames, and $1$ channel dimension. And, the generated saliency map is a matrix $\mathcal{S} \in \mathbb{R}^{H' \times W'}$. We generate post-hoc explanations using two well-known saliency methods, which produce time–frequency relevance maps that highlight regions driving model decisions. Examples of the saliency maps are shown in the first row of Figure~\ref{fig:saliencymaps}.

\textit{Occlusion Sensitivity (OS)}~\cite{zeiler2014visualizing} provides a perturbation-based explanation by systematically masking local spectrogram patches and measuring their effect on the predicted probability. Importance is attributed to regions where masking causes the largest prediction drop. This method highlights critical time–frequency regions that drive classification, and has previously been adapted to SER~\cite{kim2024speech}.  

\textit{Concept Relevance Propagation (CRP)}~\cite{achtibat2023attribution} extends Layer-wise Relevance Propagation (LRP).
Instead of attributing the importance of model inputs to predictions, CRP attributes the relevance of the inputs to higher-level concepts for more human-understandable attributions.
After defining a model concept, relevance is backpropagated through the network with respect to this concept; unrelated neurons to the selected concept are masked during explanation, yielding saliency maps that highlight the input regions most relevant for a specific concept.

\noindent\textbf{Saliency Map Segmentation:} To localize the most informative temporal regions of a saliency map, we analyze the generated saliency maps using a \textit{sliding-window} approach. We slide a fixed time-duration window across the time axis, summing the relevance values within each segment. Windows are ranked by their cumulative relevance, and the top-$k$ segments are selected in descending order. Each segment is stored with its corresponding spectrogram slice and time indices, as illustrated in the second row of Figure~\ref{fig:saliencymaps}. This procedure ensures that the most salient temporal intervals contributing to the model’s decision are retained for analysis.  

\noindent\textbf{Expert-referenced Speech Emotions Acoustics:} 
Expert-referenced acoustic cues act as theory-driven indicators of emotional expression in speech. Table~\ref{tab:vocal_cues_emotion} lists a minimalistic parameter set, defined by vocal emotion research~\cite{banse1996acoustic, scherer2003vocal, jacob2016speech, eyben2015geneva, wiechmann2025challenges, zhang2022towards}, capturing key affective markers (perceptually described) such as loudness, shrillness, breathiness/hoarseness, pitch level, and frequency or amplitude variations. In our framework, we extract these cues from the top salient segments identified in the previous step. Their magnitudes indicate \textit{what} the most relevant regions represent, providing a link between acoustic evidence and the emotional state. 

\section{Experiments}
\label{sec:experiments}
\subsection{Experimental setup}
\label{subsec: dataset}
We conduct experiments on two widely used emotional speech datasets. \textbf{CREMA-D}~\cite{cao2014crema} contains 7,442 clips from 91 actors (48 male, 43 female, aged 20–74) of diverse ethnic backgrounds, each speaking 12 sentences across six emotions (anger, disgust, fear, happiness, neutral, sadness). The dataset totals 5.26 hours of audio and is publicly available under a research license. CREMA-D is chosen for its speaker diversity and natural variability. \textbf{TESS}~\cite{SP2/E8H2MF_2020} contains 2,800 utterances from two actresses (ages 26 and 64), each producing 200 words across seven emotions (fear, neutral, happy, angry, disgust, surprise, and sad). The dataset totals 1.6 hours and is freely available for academic research.

\noindent\textbf{Data processing:} The speech samples are resampled in 16\,kHz and are preprocessed with silence removal and fixed to 2 seconds by truncation or zero-padding. We extract log-Mel spectrograms (128 filters, 1024 FFT bins) 480-sample windows, and 240-sample hops, yielding overlapping frames with minimal loss. Datasets are split into 90\% training/validation and 10\% testing with balanced emotion classes.

\noindent\textbf{Black-box SER Model:} We employ a ResNet model~\cite{he2016deep} for single-channel log-Mel spectrograms. The model has two residual blocks: the first with 32 feature maps and the second with 64, each using $3\times3$ convolutions, batch normalization, ReLU activation, and skip connections. A $2\times2$ max-pooling layer follows the first block, and global average pooling precedes the final fully connected softmax layer. Dropout (0.5) is applied before the output. Training uses Adam (batch size 64, learning rate $\{1e$-3, $5e$-4, $1e$-4$\}$), a scheduler that decreases the learning rate upon plateauing validation loss (factor 0.4, patience 4). Each experiment is repeated three times with different random seeds, and the best validation model is selected. The SER models achieved $0.63$ accuracy on \textsc{Crema-D} (validation/test) and $0.996$ (validation) and $0.990$ (test) on \textsc{TESS}.

\begingroup
\setlength{\abovecaptionskip}{1.5pt}  
\setlength{\belowcaptionskip}{0pt}  
\setlength{\textfloatsep}{4pt}      
\setlength{\floatsep}{4pt}          
\setlength{\intextsep}{4pt}         
\begin{table*}[h!]
  \centering
  \small
  \setlength{\tabcolsep}{8pt} 
  \renewcommand{\arraystretch}{.9} 
  \begin{threeparttable}[t]
  \caption{Acoustic feature statistics (mean $\pm$ std) by emotion for \textbf{True} model predictions.}
  \label{tab:cremad}
  \begin{tabular}{l l r r r r r r}
  \toprule
  \cmidrule(lr){1-2}\cmidrule(lr){3-8}
  \textbf{Emotion} & \textbf{Method} &
  \textbf{Loud.} &
  \textbf{Amp. var} &
  \textbf{Freq. var $\times 10^{-4}$} &
  \textbf{Pitch lvl.} &
  \textbf{Breath.} &
  \textbf{Shrill.(-) $\times 10^{-2}$}\\
  \midrule
Angry   & CRP & \best{2.03} $\pm$ 1.56 & 0.59 $\pm$ 0.78 & 71.00 $\pm$ 117.00 & 22.99 $\pm$ 16.68 & 2.34 $\pm$ 3.45 & \best{1.66} $\pm$ 1.55 \\
        & OS  & 1.52 $\pm$ 0.99 & \best{0.96} $\pm$ 1.08 & \best{110.00} $\pm$ 146.00 & \best{26.74} $\pm$ 14.46 & \best{3.29} $\pm$ 3.73 & 2.31 $\pm$ 1.70 \\
        \addlinespace[1.5pt]
Fear    & CRP & 0.32 $\pm$ 0.10 & 0.58 $\pm$ 0.86 & 84.00 $\pm$ 146.00 & 14.49 $\pm$ 18.06 & 2.94 $\pm$ 3.82 & \best{1.32} $\pm$ 2.22 \\
        & OS  & \best{0.48} $\pm$ 0.18 & \best{0.74} $\pm$ 0.70 & \best{120.00} $\pm$ 193.00 & \best{26.82} $\pm$ 16.45 & \best{6.01} $\pm$ 4.01 & 1.56 $\pm$ 1.89 \\
        \addlinespace[1.5pt]
Happy   & CRP & \best{0.36} $\pm$ 0.20 & \best{0.77} $\pm$ 0.86 & \best{113.00} $\pm$ 168.00 & \best{20.44} $\pm$ 19.36 & \best{4.60} $\pm$ 5.22 & \best{1.12} $\pm$ 1.50 \\
        & OS  & 0.34 $\pm$ 0.16 & 0.54 $\pm$ 1.11 & 60.00 $\pm$ 124.00 & 11.87 $\pm$ 17.73 & 1.89 $\pm$ 2.94 & 1.49 $\pm$ 1.46 \\
        \addlinespace[1.5pt]
Neutral & CRP & \best{0.35} $\pm$ 0.20 & \best{0.60} $\pm$ 0.83 & \best{59.00} $\pm$ 85.00 & \best{14.56} $\pm$ 12.73 & \best{2.75} $\pm$ 2.98 & \best{2.39} $\pm$ 1.39 \\
        & OS  & 0.23 $\pm$ 0.19 & 0.12 $\pm$ 0.43 & 16.00 $\pm$ 57.00 & 4.59 $\pm$ 10.25 & 0.73 $\pm$ 1.70 & 1.38 $\pm$ 1.08 \\
        \addlinespace[1.5pt]
Sad     & CRP & 0.19 $\pm$ 0.05 & 0.09 $\pm$ 0.28 & 9.00 $\pm$ 29.00 & 3.80 $\pm$ 10.75 & 0.78 $\pm$ 2.24 & \best{1.30} $\pm$ 2.04 \\
        & OS  & 0.19 $\pm$ 0.04 & \best{0.19} $\pm$ 0.42 & \best{31.00} $\pm$ 75.00 & \best{6.12} $\pm$ 12.81 & \best{1.25} $\pm$ 2.63 & 1.08 $\pm$ 1.11 \\
  \bottomrule
  \end{tabular}
  \begin{tablenotes}[flushleft]\footnotesize
    \item \noindent\textit{Notes:} Loud. = Loudness [sones]; Amp. var. = Shimmer [dB]; Freq. var. = Jitter [ratio]; Pitch lvl. = F0 [st, ref. 27.5 Hz]; Breath = HNR [dB]; Shrill. = spectral slope 500–1500 Hz [dB/kHz]. Values in Freq. var are scaled by $10^{-4}$ and Shrill by $10^{-2}$ for readability. For high-arousal emotions (anger, fear), higher values indicate stronger cues, except for Shrillness (more negative = stronger). Opposite holds for low-arousal emotions, consistent with Table~\ref{tab:vocal_cues_emotion}.
  \end{tablenotes}
  \end{threeparttable}
\end{table*}
\endgroup

\noindent\textbf{Saliency Segmentation:}
We compare two post-hoc saliency methods to evaluate our framework. First, \textit{OS} as in~\cite{kim2024speech}. We applied a sliding occlusion window (size 10 frames, stride 3) mask local spectrogram patches, and importance is attributed by measuring the change in class probability. The resulting map highlights critical time–frequency regions. Second, we adapted \textit{CRP} to our ResNet-based SER model using the \texttt{zennit} \cite{anders2023software} framework with the \texttt{EpsilonPlusFlat} composite and \texttt{ResNetCanonizer}. Saliency maps were then computed by conditioning relevance on either target classes (e.g., \textit{angry}, \textit{sad}), producing temporally localized saliency maps over spectrograms. Finally, for each generated saliency map, we extracted the salient regions using a fixed time duration (0.15\,s) and selected the top-$k$ windows ($k=5$) in descending order.


 \subsection{Extracting Expert-Referenced Acoustics}
\label{subsec:acoustic}
We extract the expert-referenced features that are validated in the vocal-emotion literature (Table~\ref{tab:vocal_cues_emotion}): loudness, shrillness, breathiness/hoarseness, pitch level, frequency, and amplitude variations. Features were computed using audio windows corresponding to the time indices of the previously identified salient segments using \textsc{OpenSmile}~\cite{eyben2015geneva}. We aggregated cue means with standard deviations ($\pm$SD) across windows.

\subsection{Validation Study}
\label{subsec: valid}
To assess the reliability of our framework, we designed a two-level validation procedure. For each utterance, the top-$k$ salient windows (e.g., $k=5$) were extracted and compared against two baseline metrics: (i) \textbf{full clip}, calculated from the corresponding speech waveform, and (ii) \textbf{random regions}, randomly sampled top-$k$ segments matched in time duration to the salient intervals from the same utterance.

Expert-referenced acoustic cues were then computed separately for all groups, and the average values were aggregated across windows of all groups. For each acoustic cue $f$, we computed the difference between the salient and baseline groups as follows:
\begin{equation}
\Delta f = \mu_{\text{salient}}(f) - \mu_{\text{baseline}}(f)
\end{equation}
where $\mu_{\text{salient}}(f)$ and $\mu_{\text{baseline}}(f)$ denote the mean values of cue $f$ over salient and baseline windows, respectively. As shown in Table~\ref{tab:vocal_cues_emotion}, a positive $\Delta f$ for high-arousal emotions (e.g., \textit{angry}) indicates that salient regions capture stronger magnitudes of the expected cues (e.g., loudness, pitch level). Conversely, a negative $\Delta f$ suggests that the saliency map might fail to capture the expected high-arousal cues. For low-arousal emotions (e.g., \textit{sad}), the opposite holds: salient regions are expected to align with weaker acoustic patterns. This validation provides a systematic way to assess the faithfulness of explanations against expert-referenced acoustics.

\section{Results and Discussion}
\subsection{Expert-Referenced Results on Crema-D}
Table~\ref{tab:cremad} compares feature values from CRP and OS saliency map segments for the expert-referenced features. We investigate how important features vary between high-arousal and low-arousal emotions on the \textsc{Crema-D} dataset for the \textbf{correct} model predictions. Clear acoustic patterns emerge between high- and low-arousal emotions. 
For example, \textit{angry} shows the highest loudness (CRP: $2.03\pm1.56$) and amplitude variability (CRP: $0.59\pm0.78$), consistent with its high-intensity and energetically unstable profile. 
In contrast, \textit{sad} exhibits much lower loudness (CRP: $0.19\pm0.05$) and amplitude variability (CRP: $0.095\pm0.276$), reflecting weak and softer phonation. 
Pitch level further separates the two emotions: \textit{angry} ($22.99\pm16.68$ st $\approx$ 104 Hz) is higher and more variable, indicating expressive intonation, whereas \textit{sad} ($3.80\pm10.75$ st $\approx$ 29 Hz) is much lower and flatter, consistent with a monotonous delivery. 

The results show that high-arousal states like \textit{angry} carry greater intra-emotional variability (e.g., pitch and amplitude dynamics), while low-arousal states like \textit{sad} remain more stable but subdued. 
From a psychoacoustic perspective, the sad voice appears quieter, darker (with a negative spectral slope), and breathier (lower HNR), aligning with low arousal.
In contrast, the angry voice is louder, brighter, and more irregular, matching the perception of high arousal and tension. 
Notably, CRP captures these known acoustic markers more consistently across emotions, reinforcing the framework’s ability to identify discriminative cues.

\subsection{Explanation Plausibility} We assess whether explanations remain meaningful when the model misclassifies. In such cases, the highlighted cues should contradict expert-referenced correlations
For example, in Table~\ref{tab:tess}, \textit{loudness} fails to separate emotions, with \textit{angry} nearly equal to \textit{sad}, while \textit{amplitude variation} is paradoxically higher for \textit{sad} than for \textit{angry}. Similar inconsistencies appear across cues, illustrating reduced plausibility under misclassification. In contrast, when the model correctly classifies samples on \textsc{TESS}, the cues align more clearly with expectations—for example, \textit{fear}, which shows the highest mean loudness (CRP: $1.3$), whereas \textit{sad} shows much lower mean loudness (CRP: $0.11$).
\vspace{-3mm} 
\begingroup
\setlength{\abovecaptionskip}{1.5pt}  
\setlength{\belowcaptionskip}{0pt}  
\setlength{\textfloatsep}{4pt}      
\setlength{\floatsep}{4pt}          
\setlength{\intextsep}{4pt}         
\begin{table}[htbp]
  \centering
  \footnotesize
  \setlength{\tabcolsep}{3.2pt}
  \renewcommand{\arraystretch}{1.02}
  \caption{Acoustic feature statistics for \textbf{Incorrect} predictions.}
  \label{tab:tess}
  \begin{threeparttable}[t]
  \resizebox{\columnwidth}{!}{%
    \begin{tabular}{l l r r r r r r}
    \toprule
    \cmidrule(lr){1-2}\cmidrule(lr){3-8}
    \textbf{Emotion} & \textbf{Method} &
    \textbf{Loud.} &
    \textbf{Amp. var} &
    \textbf{Freq. var} &
    \textbf{Pitch lvl.} &
    \textbf{Breath.} &
    \textbf{Shrill.(-)}\\
    \midrule
    Angry   & CRP & \best{0.45} & 0.40 & 0.009 & \best{39.33} & \best{14.35} & 0.033 \\
            & OS  & 0.28 & \best{0.41} & \best{0.011} & 15.99 & 4.68 & \best{0.019} \\
    Fear    & CRP & \best{0.30} & \best{0.22} & \best{0.009} & \best{29.32} & \best{11.23} & 0.032 \\
            & OS  & 0.15 & 0.20 & 0.003 & 10.03 & 3.37 & \best{0.010} \\
    Happy   & CRP & \best{0.40} & \best{0.61} & \best{0.060} & \best{26.48} & \best{4.21} & 0.012 \\
            & OS  & 0.22 & 0.46 & 0.003 & 9.70 & 2.27 & \best{0.008} \\
    Neutral & CRP & \best{0.22} & \best{0.50} & \best{0.012} & \best{28.22} & \best{8.90} & 0.016 \\
            & OS  & 0.16 & 0.15 & 0.003 & 10.62 & 2.64 & \best{0.006} \\
    Sad     & CRP & \best{0.43} & 0.53 & \best{0.036} & \best{34.70} & \best{6.07} & 0.011 \\
            & OS  & \best{0.40} & \best{0.56} & 0.017 & 32.74 & 4.41 & \best{0.009} \\
    \bottomrule
    \end{tabular}%
  }
  \begin{tablenotes}[flushleft]\footnotesize
    \item \parbox{\columnwidth} {Notes: Cues units as Table~\ref{tab:cremad}. Equal or higher cue values for low-arousal emotions, compared to high-arousal ones in \textit{model misclassifications}, contradict established vocal emotion theory.}
  \end{tablenotes}
  \end{threeparttable}
\end{table}
\endgroup
\vspace{-6mm} 
\subsection{Validation results}
We validated the framework using the procedure in subsection~\ref{subsec: valid}. As shown in Table \ref{tab:diff_means}, the instance-wise differences $\Delta f$ between salient regions (via CRP) and the corresponding \textbf{full clip} baseline on \textsc{Crema-D} generally aligned with expert-referenced expectations. For instance, \textit{angry} yielded positive $\Delta f$,
indicating stronger cue magnitudes in salient regions compared to the baseline, and higher than those for \textit{sad}. In contrast, low-arousal emotions such as \textit{sad} showed negative or lower $\Delta f$, consistent with their weaker acoustic profiles. Additionally, the results on \textbf{random regions} show a positive $\Delta f$ for \textit{angry} (CRP, Freq. Variation: 0.01), in contrast to a negative $\Delta f$ for \textit{sad} (CRP, Freq. Variation.: -0.01), emphasizing that the framework captures meaningful emotion-related cues rather than arbitrary patterns, demonstrating its robustness and reliability.
\begingroup
\setlength{\abovecaptionskip}{1.5pt}  
\setlength{\belowcaptionskip}{0pt}  
\setlength{\textfloatsep}{4pt}      
\setlength{\floatsep}{4pt}          
\setlength{\intextsep}{4pt}         
\begin{table}[t]
  \centering
  \scriptsize
  \setlength{\tabcolsep}{2.9pt} 
  \renewcommand{\arraystretch}{1.2}
  \begin{threeparttable}
  \caption{Validation differences  $\Delta f$ between \textbf{full clip} baseline vs.\ salient regions ( via CRP) on \textsc{Crema-D}.}
  \label{tab:diff_means}
  \begin{tabular}{l l r r r r r r}
  \toprule
  \cmidrule(lr){1-2}\cmidrule(lr){3-8}
  \textbf{Emotion} & \textbf{Method} &
  \textbf{Loud.} &
  \textbf{Amp. var} &
  \textbf{Freq. var} &
  \textbf{Pitch lvl.} &
  \textbf{Breath.} &
  \textbf{Shrill.(-)}\\
  \midrule
  Angry   & CRP &  \best{0.43} &  0.00 &  \best{0.0011} &  $-$2.55 &  $-$3.00 &  \best{$-$0.0094} \\
  Fear    & CRP & \best{0.02} &  \best{0.23} &  \best{0.0024} &  \best{3.25} &  \best{0.40} &  0.0072 \\
  Happy   & CRP & \best{0.06} &  \best{0.23} &  \best{0.0023} &  \best{6.79} &  \best{1.63} &  $-$0.0008 \\

  Neutral & CRP &  0.00 &\best{$-$0.02 }& 0.0009 &  1.96 &  0.90 & $-$0.0001 \\
  Sad     & CRP &  0.03 & \best{$-$0.045 }&\best{$-$0.0011} & \best{$-$0.69} & \best{$-$0.12} & \best{0.0040} \\
  \bottomrule
  \end{tabular}
  \begin{tablenotes}[flushleft]\footnotesize
    \item \parbox{\columnwidth} {Notes: : Cue units follow Table~\ref{tab:cremad}. Bold values reveal the higher averaged values of the segments selected by the XAI saliency regions compared to the averaged values of the corresponding full clip for high arousal emotions, and the opposite for low arousal emotions.}
  \end{tablenotes}
  \end{threeparttable}
\end{table}
\endgroup
\section{Conclusion}
In SER, the reliability of saliency-based explanations remains unexplored, limiting their trustworthiness. To address this, we introduced a novel framework that improves saliency map explanation with expert-referenced acoustic cues, clarifying \textit{``what"} features are highlighted, \textit{``why"} they matter for the model reasoning, and \textit{``where"} they occur in time.
Our results show that high-arousal states, such as \textit{angry}, exhibit stronger and more variable cues (e.g., pitch, loudness). In contrast, low-arousal states, such as \textit{sad}, display weaker. These patterns demonstrate that the framework captures emotion-relevant markers rather than random fluctuations. Moreover, we found that misclassified samples often produce implausible cue alignments, while correctly classified cases yield consistent psychoacoustic profiles, reinforcing explanation reliability. 
Future work will extend the framework to enhance concept-based XAI for speech event detection through well-identified concepts. It also enhances XAI for speech paralinguistic tasks such as pathology detection, social signal processing, and affective health monitoring.
\vfill\pagebreak
\newpage
\section{Acknowledgments}
This work is associated with the Transregional Collaborative Research Centre (TRR) 318 “Constructing Explainability” and funded by Bielefeld University.
\section{Ethical Compliance Statement}
This study uses only publicly available datasets in compliance with their licenses. No new data collection involving human participants or animals was conducted.
\\
\textbf{Source Code:} {\url{https://github.com/Seham-Nasr/BeyondSaliency-X_SER.git}}
\begin{spacing}{0.6}
\bibliographystyle{IEEEbib}
\bibliography{refs}
\end{spacing}


\end{document}